\pdfoutput=1
\documentclass[11pt]{article}

\usepackage[margin=1in]{geometry}
\usepackage{amsmath}
\usepackage{amssymb}
\usepackage{graphicx}
\usepackage{xurl}
\usepackage{caption}
\usepackage{authblk}
\usepackage[hidelinks]{hyperref}

\graphicspath{{figs/}}

\title{Lightweight Image Classification of Raptor Species for Edge
Devices: Rare-Species Dataset Expansion via Video Frame Extraction,
Knowledge Distillation, and TensorRT Deployment}

\author[ ]{Takeshi Nishikawa}
\affil[ ]{Foundation for Computational Science (FOCUS), Computational Science
Center Building 1F, 7--1--28 Minatojima-Minamimachi, Chuo-ku, Kobe
650--0047, Japan\\ \texttt{nishikawa@j-focus.or.jp}}
\date{}

\begin{document}
\maketitle

\begin{abstract}
We investigate lightweight raptor-species classification for real-time
edge deployment in wind-turbine collision mitigation. Using DINOv2-L
(304M parameters) as a teacher, we distilled three lightweight students
(MobileNetV4, ViT-Small, and EfficientNet-B0). To reduce confusion
between closely related species, we expanded the dataset to 12,519
images, including an increase in Steller's Sea Eagle images from 463 to
2,050 via video-frame extraction. Under a group split that separates
samples at the video- and source-image level to mitigate source leakage
at that granularity, the three-student ensemble achieved a macro recall
of $0.935 \pm 0.004$ over five distillation seeds (0.955 on a
conventional image-level split, retaining 97.5\% of the teacher's macro
recall) with roughly one-eighth as many parameters. On a subset of 1,258
images disjoint from the former training images, White-tailed Eagle
recall improved by up to 38.6 percentage points, while the rate at which
it was misclassified as the Steller's Sea Eagle decreased from 61\% to
15\% of errors. TensorRT FP16 deployment of EfficientNet-B0 on an NVIDIA
Jetson Orin Nano achieved 3.19 ms/image including host--device transfer
(313 images/s), with 99.95\% argmax agreement with FP32. In five-seed
controlled comparisons, neither distillation (versus CE-only) nor the
change from a DINOv2-L to a DINOv3-L teacher yielded a clear
ensemble-level improvement; the primary gains stem from the dataset
expansion and teacher re-fine-tuning.
\end{abstract}


\section{Introduction}
\label{sec:intro}

Bird strikes at wind-power facilities are an international challenge for
reconciling the conservation of rare species with the diffusion of
renewable energy~\cite{aschwanden,marques,may}. In Japan, ever since a
White-tailed Eagle (\textit{Haliaeetus albicilla}) was first confirmed
killed by a wind turbine in Tomamae, Hokkaido, in February 2004,
collisions of this species have continued to be reported, chiefly during
the wintering season~\cite{shiraki2012}. A collision-avoidance system
requires real-time detection and species identification of flying
objects to enable immediate curtailment decisions. However, on-site edge
devices (e.g., NVIDIA Jetson Orin Nano) are severely constrained in
computational resources, making it difficult to deploy a high-accuracy
classifier directly.

Knowledge distillation~\cite{hinton} is a standard solution for
transferring performance from a large teacher model to a lightweight
student, and the self-supervised foundation model DINOv2~\cite{oquab}
substantially outperforms conventional ImageNet pre-training on general
visual tasks. However, distillation performance depends on both the
quality of the teacher model and its suitability to the target domain.
When data for the target species are scarce, an iterative process is
therefore required that considers (i) data expansion, (ii)
re-fine-tuning of the teacher, and (iii) re-distillation of the
students. In particular, when morphologically similar related
species coexist---such as the White-tailed Eagle and the Steller's Sea
Eagle (\textit{H.\ pelagicus})---a lack of data for the related species
is known to blur the decision boundary between the two~\cite{ueda2010}.

The contributions of this study are the following five points.

\textbf{(1) Rare-species dataset expansion via video-frame extraction
and reduction of related-species confusion:} For the Steller's Sea
Eagle, which was initially scarce (463 images), we extracted 1,000
additional images from field-recorded videos, increasing the total to
2,050 images ($4.4\times$). Only images that satisfied both the video
creator's label and visual confirmation against the identification guide
of Maki~\cite{maki2012} were retained. Together with re-fine-tuning of
the teacher, this data expansion reduced the one-directional
misclassification of the related species White-tailed Eagle as the
Steller's Sea Eagle from 61\% to 15\% of errors and greatly improved
rare-species recall. The primary accuracy improvement
of this study stems from this combination of data expansion and teacher
re-fine-tuning.

\textbf{(2) TensorRT FP16 deployment verification on a real NVIDIA Jetson
Orin Nano:} We built a TensorRT FP16 engine of the deployed student
EfficientNet-B0 (4.02M) on the Orin Nano and measured an inference
latency of 3.19 ms/image including H2D/D2H transfer (about 313 images/s,
approximately $9.4\times$ headroom against the 30 ms target; GPU compute alone is
3.14 ms/image, 318 images/s). FP16 matched FP32 with a 99.95\% argmax
agreement over all samples, with no substantial accuracy drop observed;
no INT8 route (PTQ / QAT / quantization-friendly modification)
simultaneously exceeded FP16 in both speed and accuracy, so we fixed the
deployment at FP16.

\textbf{(3) Reaching about 97\% of the teacher's accuracy with a
lightweight student ensemble:} We re-fine-tuned the teacher DINOv2-L for
30 epochs on the 8,761-image expanded set (new-test accuracy
$0.8985\rightarrow0.9777$) and distilled its knowledge into MNV4,
ViT-Small, and EffNet-B0. The three-student ensemble reproduced a macro
recall of $0.935 \pm 0.004$ under a group split that mitigates source
leakage at the video/source-image level (five distillation seeds; 0.955
on the conventional image-level split, 97.5\% of the teacher's 0.979)
using a configuration with approximately $1/8.9$ of the teacher's
parameters.

\textbf{(4) Controlled comparison of distillation vs.\ non-distillation
(five seeds each):} Comparing distillation and non-distillation
(hard-label CE only) under identical conditions over five seeds each, the
three-student ensemble macro recall was $0.9512\pm0.0031$ (distilled) and
$0.9547\pm0.0016$ (non-distilled), with a seed-paired difference of
$-0.0034\pm0.0046$ (an exploratory comparison with $n{=}5$), whose range
includes zero and lies within seed variation. Thus, under the present
experimental setting and model scale, the additional contribution of
distillation itself at the ensemble level is limited. The primary
accuracy gains stem from data expansion and teacher re-fine-tuning.

\textbf{(5) Controlled A/B of the teacher foundation-model generation
(DINOv2-L vs.\ DINOv3-L, five seeds each):} In a controlled A/B varying
only the teacher, we evaluated the new-generation self-supervised
foundation model DINOv3-L~\cite{simeoni}, but found no clear difference
in the distilled student-ensemble accuracy ($0.9512\pm0.0031$ vs.\
$0.9533\pm0.0038$ over five seeds; seed-paired difference
$-0.0021\pm0.0063$, whose range includes zero and lies within seed
variation), showing that a foundation model should not be replaced on the
sole basis of a newer generation.

\section{Related Work and Positioning}
\label{sec:related}

\subsection{Knowledge distillation and foundation models}

Knowledge distillation~\cite{hinton} is a model-compression method that
softens a teacher model's output distribution with a temperature-scaled
softmax and trains the student jointly with a cross-entropy term; the
kinds of knowledge, teacher--student configurations, learning schemes,
and applications are organized in a survey~\cite{gou}. For Vision
Transformers in particular, DeiT~\cite{touvron} demonstrated the
effectiveness of teacher--student learning via a distillation token, and
our distillation into ViT-Small belongs to this lineage. Recently,
self-supervised foundation models typified by DINOv2~\cite{oquab} have
been reported to be effective as general-purpose visual features even in
fine-grained visual categorization (FGVC). Bird species identification is
a typical FGVC task represented by CUB-200-2011~\cite{wah} and has been
studied using large-scale, class-imbalanced real-world datasets such as
iNaturalist~\cite{vanhorn}. Using DINOv2-L as the teacher, we distill
under identical conditions into several lightweight backbones implemented
in timm~\cite{wightman} (MobileNetV4~\cite{qin}, Vision Transformer
Small~\cite{dosovitskiy}, EfficientNet-B0~\cite{tan}), and evaluate their
accuracy/model-size trade-offs under the realistic constraints of edge
devices.

\subsection{Lightweight inference and quantization for edge devices}

On on-site edge GPUs (e.g., NVIDIA Jetson Orin Nano),
TensorRT~\cite{tensorrt} accelerates inference by converting a model to
FP16 or INT8. FP16 generally yields acceleration with little accuracy
loss, while INT8 targets further acceleration and memory savings through
8-bit integer arithmetic~\cite{jacob}, but each quantization method has
its own accuracy trade-off, as organized in a survey~\cite{gholami}. In
particular, per-tensor quantization is known to cause accuracy loss in
layers with wide activation ranges; in this study, too, we experimentally
observed a large accuracy drop from PTQ of the standard EffNet-B0, which
contains Squeeze-Excite~\cite{hu} and SiLU activations. We build an FP16
engine of the deployed student on the real Orin Nano, measure its
inference latency and accuracy, and compare its trade-offs against INT8
(PTQ, quantization-aware training [QAT], and quantization-friendly
modification) on real hardware.

\subsection{Wind-turbine bird-strike countermeasures and species
identification}

In Japan, the Ministry of the Environment has published ``Guidelines for
the appropriate siting of wind-power facilities with respect to birds and
other wildlife''~\cite{moenv2011} and ``Guidelines for examining and
implementing bird-strike prevention measures for sea eagles at wind-power
facilities''~\cite{moenv2022}, yet the mechanical detection and
identification of individual birds in flight remains a technical
challenge. As representative image-based detection and species
classification of birds around wind turbines, the series of studies by
Yoshihashi et al.~\cite{yoshihashi2015,yoshihashi} stands out. They
annotated about 32,000 birds and about 4,900 non-birds with bounding
boxes and hierarchical species labels from \textbf{time-lapse images}
taken by fixed cameras installed at domestic wind-power facilities,
constructing and releasing a large dataset~\cite{yoshihashi2015}, and
evaluated the detection of relatively low-resolution small flying objects
in high-resolution wide-area images and their coarse species
classification~\cite{yoshihashi}. Our study is positioned
\textbf{downstream} of this line and differs in that (i) it specializes in
\textbf{fine-grained} identification of six raptor species cropped after
detection, and (ii) it goes as far as knowledge distillation of a
large-scale foundation model and \textbf{TensorRT deployment on real edge
hardware (Jetson Orin Nano)}. Moreover, as wind turbines grow larger
(e.g., the shift from rated 2 MW / 80 m rotor diameter to 4 MW / 150 m
class), the detection distance required for collision avoidance lengthens,
and individuals captured by cameras tend to be smaller and lower in
resolution. We therefore evaluate the classification robustness to
low-resolution inputs under PTZ magnification in Section~\ref{sec:multires}.
It has also been demonstrated that automated curtailment (automatic
shutdown of turbines at times of danger) using cameras and machine
learning can significantly reduce eagle collision
fatalities~\cite{mcclure2021,mcclure2022}, and real-time detection and
species identification of individuals in flight is a core element of
collision-avoidance systems. Ueda et al.~\cite{ueda2010} measured
differences in flight behavior between the White-tailed and Steller's Sea
Eagles and showed the higher collision risk of the former. The
bird/non-bird and IUCN-threatened-species detection stage upstream of the
species-discrimination stage targeted here is treated in a separate,
parallel study by the author (under review); this paper focuses on the
fine-grained species identification of the six protected raptors and
quantitatively reduces the classifier confusion of these two related
species through a combination of data expansion and knowledge
distillation.

\subsection{Relation to the parallel IUCN-threatened-species detection
system}
\label{sec:vs_v6}

This study differs in role and technique from a separate, parallel study
by the author (under review; title and venue given in the submission
cover sheet), ``Detecting IUCN-threatened species from bird images with
transfer-learned CNN features and a hierarchical lightweight
classifier,'' which targets the same bird-strike prevention system. That
parallel study classifies a single wild-bird image up to the IUCN
category using a five-stage hierarchical cascade of ResNet50 features and
LightGBM, serving as a primary CPU-edge screening. This paper specializes
in its downstream \textbf{fine-grained species-identification stage} for
the six protected raptors, deploying a distilled student ensemble via
TensorRT FP16 on the Orin Nano (differences in Table~\ref{tab:vs_v6}). In
a system where flying objects detected by a fixed wide-angle camera are
imaged and identified by a PTZ camera, the two work complementarily as
wide-area screening and definitive species identification.

\begin{table}[t]
\centering
\caption{Differences between the companion IUCN-detection study (under
review) and this work.}
\label{tab:vs_v6}
{\footnotesize
\begin{tabular}{p{20mm}|p{50mm}|p{50mm}}\hline\hline
Aspect & Companion study (IUCN detection) & This work (raptor species ID) \\\hline
Target / output & IUCN category of a broad range of birds (71 species $\to$ CR/EN/VU, etc.) & Species identification of six protected raptors \\\hline
Feature extraction & ResNet50 transfer learning & DINOv2-L distillation (self-supervised foundation) \\\hline
Classifier & Hierarchical cascade LightGBM & Three-student ensemble \\\hline
Deployment & CPU edge PC & Edge GPU (Orin Nano, FP16) \\\hline
\end{tabular}\par}
\end{table}

\section{Proposed Method}
\label{sec:method}

\subsection{Target species and dataset}

The targets are the six raptor species shown in Table~\ref{tab:species}.
They are distributed across four subfamilies of the Accipitriformes; in
the IUCN Red List~\cite{iucn}, only \textit{H.\ pelagicus} (Steller's Sea
Eagle) is Vulnerable, and the other five species are Least Concern.

Our dataset was built in two stages. Initially, we trained the teacher
and student models on 5,855 images across six species collected from May
2020 to November 2024 (hereafter, the \textit{old data}), but the recall
of rare species---especially the White-tailed Eagle---did not reach a
practical level (Section~\ref{sec:eval}). To resolve this shortfall, we
continued video-frame extraction and still-image acquisition from January
2025 to February 2026, expanding to 12,519 images across the six species
(hereafter, the \textit{new data}; $2.14\times$ the old data). This paper
contrasts the old model (trained on the old data) with the new model
(trained on the new data) because of this construction history, aiming to
quantify, on the same pipeline, the effect of data expansion on
rare-species classification accuracy. Part of the bird photographs and
flight videos in the old data collected from May 2020 to November 2024
were obtained with the cooperation of Eurus Energy Holdings Corporation.

\begin{table}[t]
\centering
\caption{The six target raptor species.}
\label{tab:species}
{\footnotesize
\begin{tabular}{l|l|l|c}\hline\hline
English name & Scientific name & Subfamily & IUCN \\\hline
Golden Eagle & \textit{Aquila chrysaetos} & Aquilinae & LC \\
White-tailed Eagle & \textit{H.\ albicilla} & Haliaeetinae & LC \\
Steller's Sea Eagle & \textit{H.\ pelagicus} & Haliaeetinae & VU \\
Mountain Hawk-Eagle & \textit{Nisaetus nipalensis} & Aquilinae & LC \\
Northern Goshawk & \textit{Accipiter gentilis} & Accipitrinae & LC \\
Eastern Marsh Harrier & \textit{Circus spilonotus} & Circinae & LC \\\hline
\end{tabular}\par}
\end{table}

Data were collected through the following two routes.

\noindent\textbf{(a) Video-frame extraction:}\, In this study, we applied
frame extraction from field-recorded videos to all six species, adding to
the set of video frames already prepared for training our existing
bird-image classifier (the sixth version of a bird multi-class classifier
built by our group, hereafter V6). Extraction was conditioned on a
confidence of at least 0.7 from a trained bird detector after
time-interval sampling, followed by duplicate removal with the perceptual
hash pHash (Hamming distance below 8). Because the Steller's Sea Eagle
was initially scarce at 463 images relative to other species, we
additionally extracted 1,000 images, 16--31 per video, from 33 of 35
Steller's Sea Eagle-only videos. The recording locations of the videos
were, by species: Golden Eagle in Kamaishi and Tono (Iwate) and Maibara
(Shiga); White-tailed Eagle in Wakkanai and Toyotomi (Hokkaido);
Steller's Sea Eagle in Wakkanai (Hokkaido) and Nagahama (Shiga); Mountain
Hawk-Eagle in Maibara and Higashiomi (Shiga), Kobe (Hyogo) and Himi
(Toyama); Northern Goshawk in Nagareyama and Kashiwa (Chiba) and Tsukuba,
Tsuchiura, and Inashiki (Ibaraki); and Eastern Marsh Harrier in Himeji
(Hyogo), Nagahama (Shiga), Oarai (Ibaraki) and Himi (Toyama).

\noindent\textbf{(b) Four still-image routes:}\, From four routes---
iNaturalist (via GBIF)~\cite{inat2024}, GBIF~\cite{gbif}, Wikipedia, and
Wikimedia Commons---we obtained only images carrying a redistributable
and modifiable license (one of CC BY 4.0, CC BY-SA 4.0, or CC0 1.0). For
each image, we saved the occurrence ID, photo ID, acquisition date,
source URL, license type, and attribution as a manifest, ensuring rights
handling and reproducibility (however, iNaturalist is at the photo-ID
level and does not include an observation-level ID). For duplicate
removal, both within and across routes, we first removed exact duplicates
by matching photo IDs, occurrence IDs, and source URLs, and further
removed near-duplicates from video frames and Steller's Sea Eagle
augmentation candidates using pHash (Hamming distance below 8). However,
we did not perform perceptual-hash duplicate removal across all routes, so
the same photo re-posted on different routes (e.g., iNaturalist and GBIF,
or Wikipedia and Wikimedia Commons) may remain as separate files due to
differences in IDs or URLs (future work).

Species identification used both the video creator's label (species names
in video titles and descriptions) and morphological visual confirmation
against the eagle/hawk/falcon identification guide of
Maki~\cite{maki2012}; any set with a discrepancy between the two was not
adopted. The image counts before and after expansion are shown in
Table~\ref{tab:dataset}.

\begin{figure}[t]
\centering
\includegraphics[width=\linewidth]{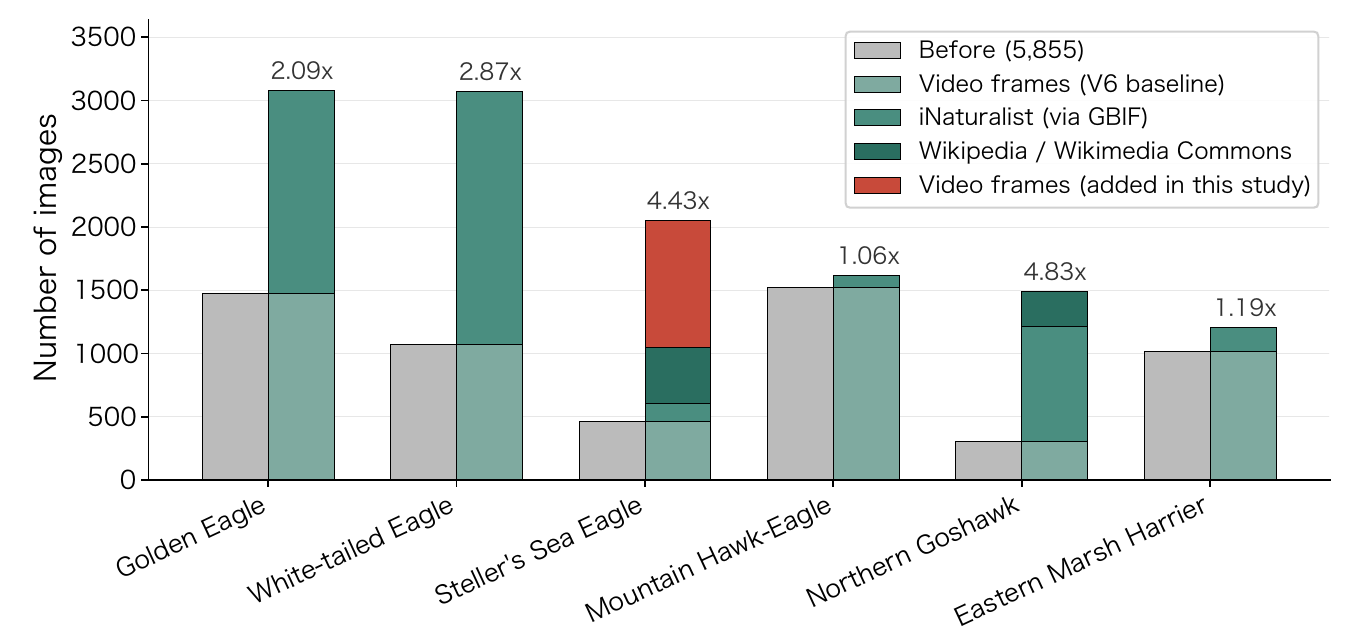}
\caption{Numbers of images per species before (gray) and after (stacked)
dataset expansion.}
\label{fig:dataset_growth}
\end{figure}

\begin{table}[t]
\centering
\caption{Number of images per species before/after dataset expansion
(12,519 in total).}
\label{tab:dataset}
{\footnotesize
\begin{tabular}{l|r|r|l}\hline\hline
Species & Before & After & Increment breakdown \\\hline
Golden Eagle & 1,477 & 3,081 & +1,604 (iNat) \\
White-tailed Eagle & 1,069 & 3,069 & +2,000 (iNat) \\
\textbf{Steller's Sea Eagle} & \textbf{463} & \textbf{2,050} & \textbf{+1,000 (video) +141 iNat +446 Wiki} \\
Mountain Hawk-Eagle & 1,522 & 1,616 & +94 (iNat) \\
Northern Goshawk & 309 & 1,494 & +909 iNat +276 Wiki \\
Eastern Marsh Harrier & 1,015 & 1,209 & +194 (iNat) \\\hline
\textbf{Total} & \textbf{5,855} & \textbf{12,519} & \textbf{+6,664 ($2.14\times$)} \\\hline
\end{tabular}\par}
\end{table}

By stratified splitting (random seed 42, ratio 0.7/0.15/0.15), we divided
the data into 8,761 training / 1,876 validation / 1,882 test images. The
pre-expansion test set ($n{=}883$) and the post-expansion test set
($n{=}1{,}882$) are different populations even under the same seed. For a
fair comparison of the old and new models, we separately constructed a
\textit{subset disjoint from the former training images} ($n{=}1{,}258$)
by removing from the new test set the samples that overlap with the old
training set.

\subsection{Species discrimination in additionally collected data}
\label{sec:gemini}
The videos used for frame extraction were those recorded by the author in
the field or those for which usage permission was obtained from the
videographer, all limited to rights-cleared material (recording locations
given in the previous section). On the other hand, for videos on
video-sharing services referred to only to assist the primary assignment
of species labels, we did not directly obtain or store the video files;
instead, we input the video URL to a multimodal large-language-model API
(Google Gemini API~\cite{gemini}, model gemini-2.5-pro, executed February
2026) and obtained species candidates via multimodal inference over the
video content on the API server. The resulting discriminations were
combined with the video creator's label and visual confirmation against
the identification guide of Maki~\cite{maki2012} to form a primary species
label; any set with a discrepancy was not adopted. A prompt constraint
that prohibited speculative judgments against backgrounds of only sky or
clouds and returned \texttt{NONE} when no bird was present suppressed
false positives.

\subsection{Re-fine-tuning of the teacher DINOv2-L}

\begin{figure}[t]
\centering
\includegraphics[width=\linewidth]{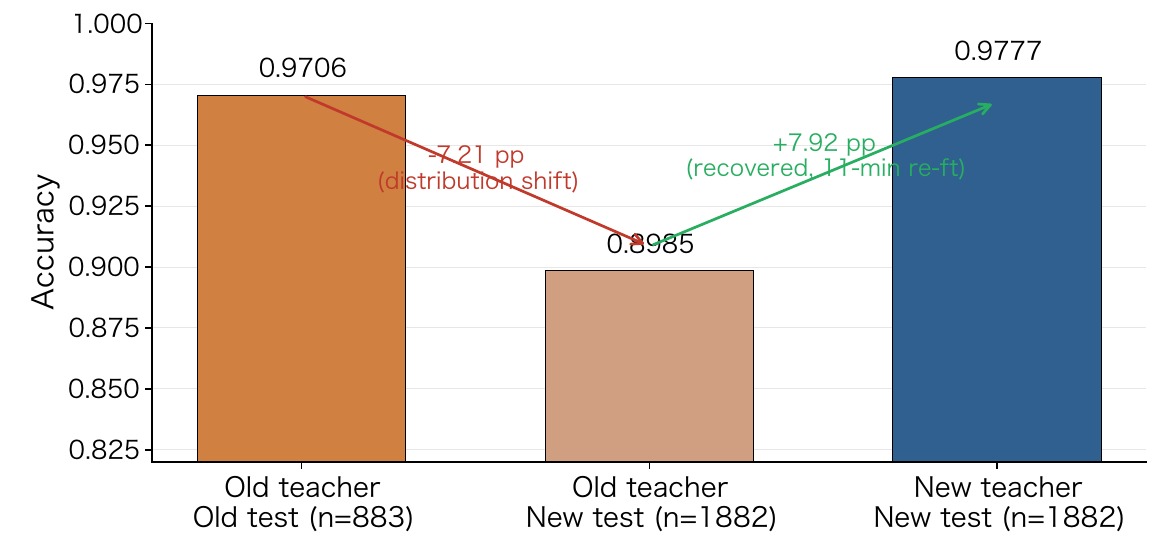}
\caption{Performance change of the DINOv2-L teacher under dataset
expansion and recovery by re-fine-tuning.}
\label{fig:teacher_recovery}
\end{figure}

For the teacher model, we used DINOv2 ViT-L/14~\cite{oquab,dosovitskiy}:
after loading pre-trained weights via timm~\cite{wightman}, we added a
classification head for the six-class problem. We used $224\times224$
pixel input, AdamW (backbone learning rate $1{\times}10^{-5}$, head
learning rate $1{\times}10^{-3}$, weight decay 0.05), batch 64, two-GPU
parallelism via DDP~\cite{torchddp}, 30 epochs, patience 10, cosine
learning-rate decay, and augmentation with random resized crop,
horizontal flip, and color jitter. The teacher re-fine-tuned on the
expanded training set achieved an accuracy of 0.9777 and a macro recall of
0.9789 on the new test set (+7.92 points over the old teacher's accuracy of
0.8985 on the new test set; see Figure~\ref{fig:teacher_recovery}).

\subsection{Knowledge distillation of the three students}

The students are the three models shown in Table~\ref{tab:students}, built
with timm (hereafter, MobileNetV4 is abbreviated as MNV4 and
EfficientNet-B0 as EffNet-B0). The loss function is the linear sum of the
following temperature-scaled distillation loss~\cite{hinton} and
hard-label cross-entropy.
\begin{align}
\mathcal{L} &= \alpha\, T^2\, \mathrm{KL}\!\left(\mathrm{softmax}(t/T)\,\|\,\mathrm{softmax}(s/T)\right) \nonumber\\
&\quad + (1-\alpha)\, \mathrm{CE}(s, y)
\label{eq:kd}
\end{align}
where $s$ and $t$ are the student and teacher logits, respectively. The
first term is the forward KL divergence $\mathrm{KL}(p_t\,\|\,p_s)$
targeting the teacher distribution, corresponding in implementation to
\texttt{F.kl\_div(} \texttt{log\_softmax($s$/T),} \texttt{softmax($t$/T),}
\texttt{reduction="batchmean")} (normalized by batch mean). All students were
initialized from ImageNet-pretrained weights; the exact timm identifiers,
parameter counts, distillation temperatures $T$, and weights $\alpha$ are
given in Table~\ref{tab:students}. $T$ and $\alpha$ were selected for each
student by a preliminary sweep on the validation set (the test set was
used only for final evaluation). Training used batch 256, AdamW (backbone
$5{\times}10^{-4}$, head $5{\times}10^{-3}$, weight decay 0.05), and 80
epochs (patience 10, EffNet-B0/MNV4).

\begin{table}[t]
\centering
\caption{The three student models (timm identifier, parameter count,
distillation $T$, $\alpha$); all initialized from ImageNet-pretrained
weights.}
\label{tab:students}
{\footnotesize
\begin{tabular}{l|l|r|c|c}\hline\hline
Model & timm identifier & Params & $T$ & $\alpha$ \\\hline
MNV4 & \texttt{mobilenetv4\_conv\_medium.e500\_r256\_in1k} & 8.44M & 4.0 & 0.7 \\
ViT-Small & \texttt{vit\_small\_patch16\_224.augreg\_in21k\_ft\_in1k} & 21.67M & 4.0 & 0.7 \\
EffNet-B0 & \texttt{efficientnet\_b0.ra\_in1k} & 4.02M & 2.0 & 0.5 \\\hline
\end{tabular}\par}
\end{table}

\subsection{Ensemble}

The final configuration ensembles the outputs of the three students MNV4,
ViT-Small, and EffNet-B0 by averaging softmax probabilities. All
hyperparameters, including $T$, $\alpha$, early stopping, and best
checkpoint, were determined using only the validation set; the test set
was used only for final evaluation.

\subsection{TensorRT FP16 deployment on the Orin Nano}
\label{sec:deploy_method}

We exported the best.pt of the deployed student EffNet-B0 (4.02M) to ONNX
(opset 17, no dynamic axes, fixed batch 1)~\cite{onnx} and built an FP16
engine (10.0 MB) with \texttt{trtexec --fp16} on an NVIDIA Jetson Orin
Nano (JetPack 5.1.2 / TensorRT 8.5.2)~\cite{tensorrt}. Latency was
measured under an exclusive condition (with concurrent processes
excluded) with \texttt{--warmUp=1000 --iterations=1000 --avgRuns=100
--useSpinWait}. FP16 accuracy was verified by comparing the argmax of
predictions passed through the real engine with the FP32 predictions of
onnxruntime on a Mac. For comparison, INT8 was verified on real hardware
via three routes: (a) PTQ with \texttt{libnvinfer} 8.5.2's
\texttt{IInt8EntropyCalibrator2} calibrated on 200 training
images~\cite{jacob}; (b) QAT (quantization-aware training) with an Q/DQ
engine via the NVIDIA TensorRT Model Optimizer~\cite{modelopt}; and (c)
PTQ of a quantization-friendly modified version with Squeeze-Excite
removed and SiLU$\to$ReLU6 substituted.

\section{Evaluation Experiments}
\label{sec:eval}

\subsection{Experimental setup}

The primary metric is \textbf{macro recall}; as secondary metrics we
report accuracy, per-class recall, and confusion matrices, and the
operational precision metrics (macro precision, aggregated precision, and
false-stop rate) are reported in Section~\ref{sec:stopdecision}. Because
in a collision-avoidance system the cost of a miss exceeds the cost of a
false positive (an erroneous shutdown), we chose macro recall as the
primary metric. As computing resources, we used the FOCUS supercomputer
M system, HPCI-AISS1, and an NVIDIA Jetson Orin Nano (fgpu0) of the
Foundation for Computational Science, and used a MacBook Pro (Apple M2
Max) for development and model conversion. The specifications of each
system are given in Table~\ref{tab:hw}.

\begin{table}[t]
\centering
\caption{Computing resources used in this study.}
\label{tab:hw}
{\footnotesize
\begin{tabular}{p{30mm}|p{62mm}|p{34mm}}\hline\hline
System & Key specifications & Use in this study \\\hline
FOCUS M system~\cite{focus} & Xeon Platinum 8480+ $\times$2, 1.0 TiB, MN-Core 2 $\times$8 & Data preprocessing / CPU-side inference \\\hline
HPCI-AISS1~\cite{h200} & Xeon 6520P $\times$2, 1.5 TiB, H200 NVL $\times$4 (560 GiB total) & Teacher/student training, QAT \\\hline
Jetson Orin Nano 8 GB~\cite{orin} & Ampere GPU (1,024 CUDA cores), up to 40 TOPS, 7--15 W & Edge inference latency \& accuracy measurement \\\hline
MacBook Pro (M2 Max)~\cite{m2max} & Apple M2 Max, unified 64 GB & ONNX conversion / FP32 reference \\\hline
\end{tabular}\par}
\end{table}

\subsection{Distillation results of the three students}

Table~\ref{tab:students_acc} shows the standalone accuracy of the three
students on the new test set ($n{=}1{,}882$). The v3 distillation using the
new-teacher logits exceeded the old-teacher-logit version (v2) for all
students, with ViT-Small showing the largest improvement of +3.24 points
(values computed by the same single evaluation procedure as
Section~\ref{sec:eval}).

\begin{table}[t]
\centering
\caption{Distillation results for the three students on the new test set
($n{=}1{,}882$). $\Delta$ acc = v3 $-$ v2.}
\label{tab:students_acc}
{\footnotesize
\begin{tabular}{l|r|r|r|r}\hline\hline
Model & Params & v2 acc & v3 acc & $\Delta$ (points) \\\hline
MNV4 & 8.44M & 0.9022 & 0.9038 & +0.16 \\
\textbf{ViT-Small} & 21.67M & 0.9102 & \textbf{0.9426} & \textbf{+3.24} \\
EffNet-B0 ($T{=}2,\alpha{=}0.5$) & 4.02M & 0.9230 & 0.9283 & +0.53 \\\hline
\end{tabular}\par}
\end{table}

\subsection{Reaching about 97\% of the teacher's accuracy with a
lightweight ensemble}

By averaging the softmax outputs of the three students MNV4, ViT-Small,
and EffNet-B0, the ensemble achieved an accuracy of 0.9506 and a macro
recall of 0.9547 on the standard image-level-split test set
(Table~\ref{tab:ens}). This reproduces 97.5\% of the level of the teacher
DINOv2-L (accuracy 0.9777, macro recall 0.9789) with a configuration of
34.13M parameters in total, approximately $1/8.9$ of the teacher's
parameters. All six species maintained a recall of at least 0.93, and in
particular the recall of the IUCN VU species Steller's Sea Eagle was
0.9708. However, because this standard split may include source leakage
from frames of the same video, we regard the macro recall of $0.935 \pm
0.004$ (five seeds) under the group split separated at the
video/source-image level (Section~\ref{sec:group_split}) as the primary
measure of unseen-data generalization, and treat the 0.955 in this
section as a reference. From this section onward, all evaluation values
were computed by a single evaluation procedure using the same
preprocessing as deployment (resize to short side 256, then central
$224$ crop).

\begin{table}[t]
\centering
\caption{Accuracy of the three-student ensemble (standard image-level
split test set, $n{=}1{,}882$; reference).}
\label{tab:ens}
{\footnotesize
\begin{tabular}{l|r|r}\hline\hline
Configuration & Accuracy & Macro recall \\\hline
Teacher DINOv2-L (304M) & 0.9777 & 0.9789 \\
\textbf{Three-student ensemble $\star$} & \textbf{0.9506} & \textbf{0.9547} \\\hline
\end{tabular}\par}
\end{table}

\subsection{Effect of knowledge distillation (comparison with a
non-distillation baseline)}
\label{sec:ceonly}

To isolate the contribution of knowledge distillation itself, we built a
non-distillation baseline that fixed the split, data augmentation,
optimizer, and sampler identically for the three students but removed the
distillation signal and trained only with the weighted hard-label
cross-entropy (corresponding to $\alpha{=}0$ in Eq.~(\ref{eq:kd})). First,
the standalone accuracy at a single seed (seed 42) is shown in
Table~\ref{tab:ceonly}. For the deployment target EffNet-B0, distillation
improved macro recall from 0.9283 to 0.9369, whereas for the standalone
accuracy of ViT-Small and MNV4 the non-distilled variant was comparable to
or slightly better; thus the effect of distillation depended on the
student architecture.

To verify that this single-seed tendency was not by chance, we
independently retrained both KD and CE-only over five seeds
($\{42,\dots,46\}$) each and compared per-student and ensemble macro
recall as mean$\pm$SD (Table~\ref{tab:ceonly}, middle and bottom). For the
per-student results (middle), the deployment target EffNet-B0 gave KD
$0.9277\pm0.0059$ vs.\ CE-only $0.9261\pm0.0160$ (paired difference
$+0.0017\pm0.0156$; KD was superior in 1 of 5 seeds), so the distillation
advantage observed at single seed 42 ($+0.0086$) shrank to $+0.0017$ in
the five-seed mean and was buried within the SD. The paired differences
for ViT-Small and MNV4 were $-0.0009\pm0.0043$ and $+0.0077\pm0.0183$,
respectively, with inconsistent signs, so even for standalone students the
distillation advantage was within seed variation. For the ensemble
(bottom), KD was $0.9512\pm0.0031$ and CE-only was $0.9547\pm0.0016$. We
analyze this in two ways with an explicit evaluation target.
\textbf{(i)} The seed-paired difference (KD$-$CE) was $-0.0034\pm0.0046$
(an exploratory comparison with $n{=}5$; KD was superior only at seed 42),
and the mean$\pm$SD range included zero. \textbf{(ii)} Separately, for a
single ensemble integrating the five seeds by majority vote, an
image-level paired bootstrap ($B{=}2000$) evaluating the uncertainty of
the test sample gave a difference of $-0.0036$, a 95\% confidence interval
of $[-0.0086, +0.0011]$, and a fraction of resamples in which KD exceeded
CE of 0.068 (this fraction is not a usual $p$-value but a fraction among
bootstrap resamples). Both (i) and (ii) include zero and do not support a
clear improvement from distillation (because we did not perform a formal
seed-level test, we say ``no clear improvement was confirmed'' rather than
``significant''). \textbf{That is, for both standalone students and the
ensemble, the difference between knowledge distillation and
non-distillation was within seed variation, and no clear improvement from
distillation was confirmed.} The distillation advantage observed at a
single seed is attributable to seed variation. In this study, we adopt the
distillation configuration because distillation does not harm accuracy for
the single deployed EffNet-B0 and for design consistency in leveraging
teacher knowledge, but the primary accuracy improvement in this task stems
from data expansion and teacher re-fine-tuning
(Section~\ref{sec:eval}), and the additional contribution of distillation
itself is limited for both standalone students and the ensemble.

\begin{table}[t]
\centering
\caption{Knowledge distillation vs.\ non-distillation (CE-only) under
identical settings (new test set, $n{=}1{,}882$, macro recall). Top:
per-student at seed 42 (same models as Table~\ref{tab:students_acc});
middle: per-student mean$\pm$SD over five independently retrained seeds;
bottom: ensemble mean$\pm$SD over the same five seeds.}
\label{tab:ceonly}
{\footnotesize
\begin{tabular}{l|r|r|r}\hline\hline
Configuration & CE-only & KD & Diff \\\hline
\multicolumn{4}{l}{\footnotesize Per-student (seed 42, same models as Table~\ref{tab:students_acc})} \\
EffNet-B0 (deployed) & 0.9283 & \textbf{0.9369} & $+0.0086$ \\
ViT-Small & \textbf{0.9487} & 0.9470 & $-0.0017$ \\
MNV4 & \textbf{0.9202} & 0.9091 & $-0.0111$ \\\hline
\multicolumn{4}{l}{\footnotesize Per-student (five-seed mean$\pm$SD, independent retraining batch)} \\
EffNet-B0 (deployed) & $0.9261\pm0.0160$ & $\mathbf{0.9277\pm0.0059}$ & $+0.0017\pm0.0156$ \\
ViT-Small & $\mathbf{0.9483\pm0.0017}$ & $0.9474\pm0.0041$ & $-0.0009\pm0.0043$ \\
MNV4 & $0.9057\pm0.0201$ & $\mathbf{0.9134\pm0.0075}$ & $+0.0077\pm0.0183$ \\\hline
\multicolumn{4}{l}{\footnotesize Three-student ensemble (five-seed mean$\pm$SD)} \\
Macro recall & $\mathbf{0.9547\pm0.0016}$ & $0.9512\pm0.0031$ & $-0.0034\pm0.0046$ \\\hline
\end{tabular}\par
{\footnotesize\raggedright (ii) Image-level bootstrap of the five-seed
integrated ensemble: 95\% CI $=[-0.009,+0.001]$, KD-superior fraction
0.068 (not a $p$-value). The five seeds of the middle/bottom rows are an
independent retraining series from the top row; the seed-42 values do not
exactly match the top row.\par}}
\end{table}

\subsection{Old vs.\ new model comparison on the subset disjoint from
the former training images}

\begin{figure}[t]
\centering
\includegraphics[width=\linewidth]{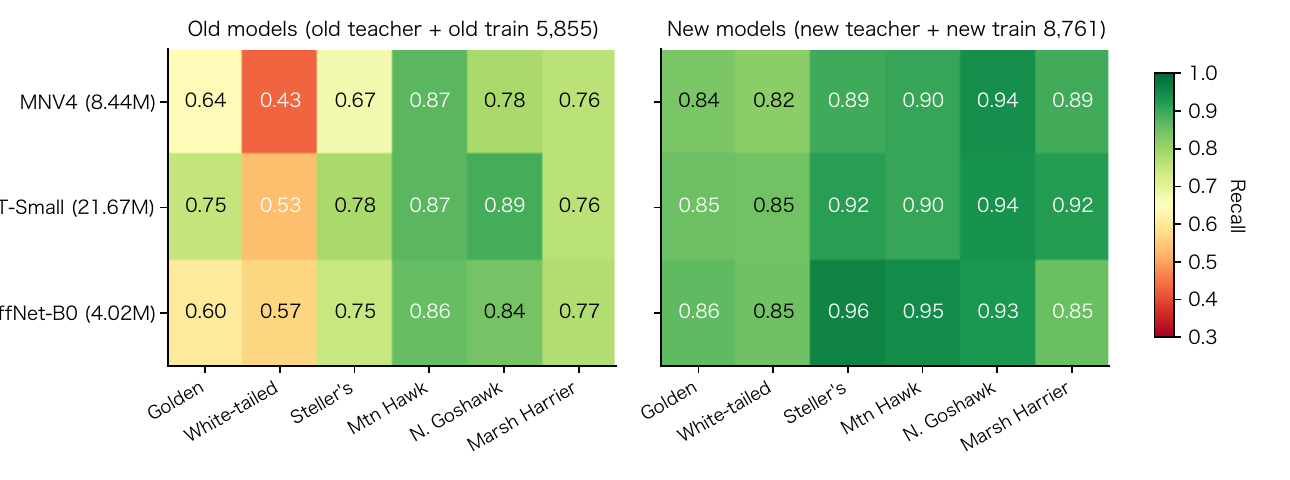}
\caption{Per-class recall on the subset disjoint from the former
training images (old left, new right).}
\label{fig:clean_subset}
\end{figure}

On the \textit{subset disjoint from the former training images} ($n{=}1{,}258$),
obtained by removing from the new test set the 624 images that were in the
old training set (determined by exact match of image relative paths with
the old training split; perceptually similar but different-path
near-duplicates are not removed), the condition is unseen for both the old
and the new models. The comparison of the three students under this fair
condition is shown in Table~\ref{tab:clean_subset} and
Figure~\ref{fig:clean_subset}.

\begin{table}[t]
\centering
\caption{Old vs new models on the subset disjoint from the former
training images ($n{=}1{,}258$).}
\label{tab:clean_subset}
{\footnotesize
\begin{tabular}{l|r|r|r|r|r}\hline\hline
Model & Old acc & New acc & $\Delta$ (pts) & Old WTE recall & New WTE recall \\\hline
EffNet-B0 & 0.6868 & 0.8935 & +20.7 & 0.5657 & 0.8486 \\
ViT-Small & 0.7273 & 0.8839 & +15.7 & 0.5343 & 0.8457 \\
MNV4 & 0.6343 & 0.8672 & +23.3 & 0.4314 & 0.8171 \\\hline
\end{tabular}\par}
{\footnotesize \hspace{1em}WTE: White-tailed Eagle.\par}
\end{table}

The White-tailed Eagle recall of 0.9255 observed on the old test set
($n{=}883$) is an overestimate due to a bias of the set (V6 series only);
under the stricter unseen condition of the subset disjoint from the
former training images, the generalization performance was confirmed to
be around 0.85.
Even so, the improvement from 0.43--0.57 (old model) to 0.82--0.85 (new
model) was at least +28 points for all three students, with the largest
being MNV4's +38.6 points ($0.4314\rightarrow0.8171$).

\subsection{Resolution of the one-directional misclassification of
White-tailed Eagle as Steller's Sea Eagle}

\begin{figure}[t]
\centering
\includegraphics[width=0.9\linewidth]{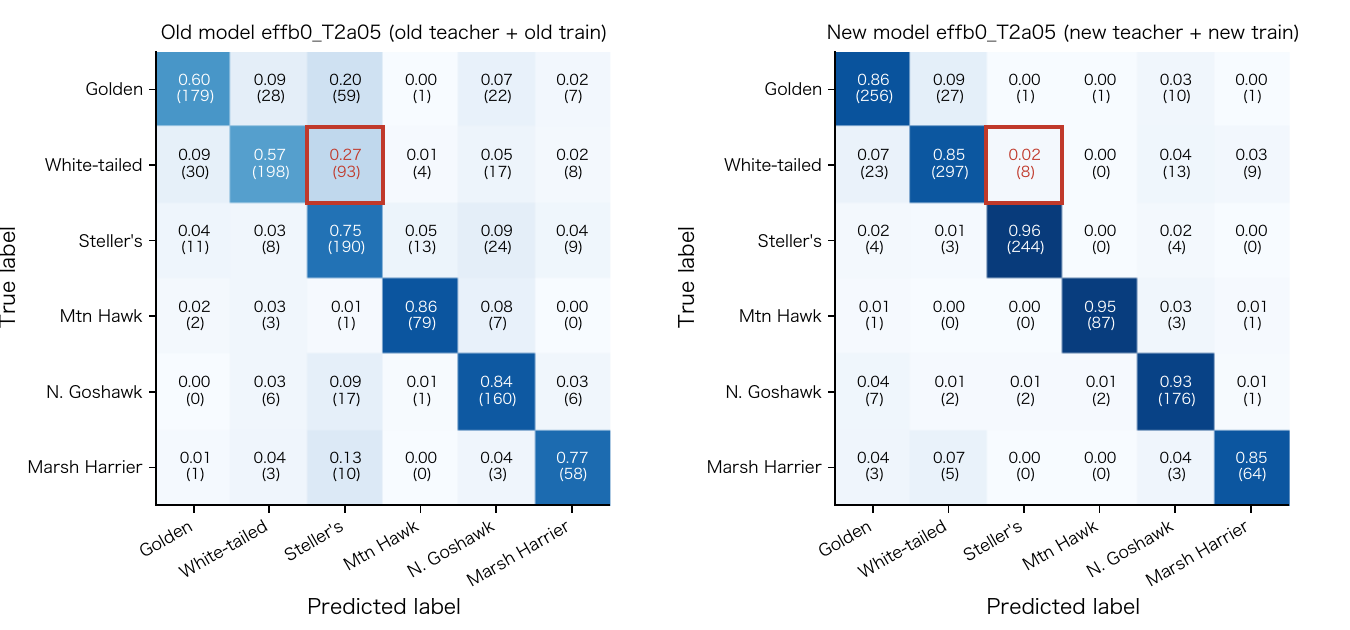}
\caption{Confusion matrices of EffNet-B0 on the subset disjoint from the
former training images (old left, new right). Red boxes indicate
misclassification of White-tailed Eagle as Steller's Sea Eagle.}
\label{fig:cm}
\end{figure}

Boundary blurring between the related species (White-tailed Eagle and
Steller's Sea Eagle, both Haliaeetinae) manifests as a bias in
one-directional misclassification. According to the confusion matrix of
EffNet-B0 (Figure~\ref{fig:cm}), in the old model 93 of the 152
misclassifications of White-tailed Eagle (61\%) were assigned to the
Steller's Sea Eagle. In the new model this decreased to 8 of 53 (15\%). Because the
new model simultaneously increases the Steller's Sea Eagle training count
from 324 to 1,432 images ($4.4\times$) and re-fine-tunes the teacher and
re-distills the students, this result is not a causal effect isolating
each factor, but is an improvement consistent with the augmentation of
Steller's Sea Eagle training images, suggesting a possible improvement of
the decision boundary for the two related Haliaeetinae species.

\subsection{Evaluation of source leakage via a group split}
\label{sec:group_split}

Because frames of the same video share the same individual, background,
and nearby poses, an image-level stratified split may spread frames of the
same video across training and test, causing source leakage to
overestimate accuracy. To verify this directly, we defined groups as (i)
video ID for video frames (33 Steller's Sea Eagle videos), (ii)
source-image level for V6-derived crops, and (iii) image level for others
(iNaturalist, GBIF, Wiki, etc.), and constructed a \textit{group split}
so that no group straddled training/validation/test (training 8,763 /
validation 1,880 / test 1,876, verified that every group is closed within
a single split). We retrained and re-evaluated under the same protocol as
the standard split (teacher re-fine-tuning, logit pre-computation,
three-student distillation); results are shown in Table~\ref{tab:group}.

The three-student ensemble macro recall on the group split (leakage
mitigated at the video/source-image level) was 0.9326, a drop of only
about 2.2 points relative to the standard image-level split (0.9547)
computed by the same evaluation procedure. The teacher likewise dropped by
1.96 points, from 0.9789 to 0.9593. All six species maintained an ensemble
recall of at least 0.915 (Steller's Sea Eagle 0.919, White-tailed Eagle
0.937), confirming that the improvement of our method is not mainly an
artifact of same-video leakage but is preserved even under conditions that
mitigate video/source-image-level leakage. Furthermore, to confirm that
this 0.9326 is not a single value dependent on the training random seed, we
retrained and re-evaluated on the same group split while varying the
student-distillation seed over five values ($\{42,\dots,46\}$). The
three-student ensemble macro recall was $0.9348\pm0.0042$ (range
0.9288--0.9384; per-species macro recall from Steller's Sea Eagle 0.909 to
Mountain Hawk-Eagle 0.957), consistent with the single-seed value and
robust to the training random seed. We regard this five-seed mean of
$0.935\pm0.004$ as the primary generalization performance. However, this
evaluation uses a single group split (split seed 42); a robustness
evaluation across split seeds is future work. Note that this group split
guarantees group separation only at the video-ID level (newly extracted
Steller's Sea Eagle video frames) and the source-image level (V6-derived
crops); grouping down to the iNaturalist observation unit (multiple photos
of the same observation) or to the frame unit of non-V6 videos cannot be
recovered from the file names, so removing leakage at that granularity is
future work.

\begin{table}[t]
\centering
\caption{Group split (leakage mitigated at video/source-image level;
primary generalization measure) vs.\ standard image-level split
(reference); macro recall under the same evaluation procedure.}
\label{tab:group}
{\footnotesize
\begin{tabular}{l|r|r|r}\hline\hline
Stage & Standard split & Group split & Diff \\\hline
Teacher DINOv2-L & 0.9789 & 0.9593 & $-0.0196$ \\
Student EffNet-B0 & 0.9369 & 0.9038 & $-0.0331$ \\
Student ViT-Small & 0.9470 & 0.9325 & $-0.0145$ \\
Student MNV4 & 0.9091 & 0.8877 & $-0.0214$ \\\hline
\textbf{Three-student ensemble} & \textbf{0.9547} & \textbf{0.9326} & $-0.0221$ \\\hline
\end{tabular}\par}
{\footnotesize \hspace{1em}The group-split column is for a single split and
single seed. Over five student-distillation seeds on the same split, the
three-student ensemble macro recall is $0.9348\pm0.0042$.\par}
\end{table}

\subsection{TensorRT FP16 deployment on the real Orin Nano}
\label{sec:orin_fp16}

We built an FP16 engine of the deployed student EffNet-B0 (4.02M) on an
NVIDIA Jetson Orin Nano (JetPack 5.1.2 / TensorRT 8.5.2, power mode 15 W
MAXN) and measured latency on real hardware; results are shown in
Table~\ref{tab:orin_latency}. The mean GPU compute was 3.139 ms/image (318
images/s), and even the TensorRT inference latency including H2D/D2H
transfer was 3.192 ms/image (about 313 images/s). These figures are all
\textbf{model inference latencies} (including H2D/D2H transfer) and
exclude preprocessing steps such as image decoding, resizing, central
cropping, and normalization; they are therefore not end-to-end
latencies. The target of 30 ms/image is set as the time budget for
processing one crop in synchronization with the standard processing rate
(about 30 fps) of the detection/tracking stage, assuming an operation that
identifies a flying object tracked by a PTZ camera on every frame. Against
this target, the inference latency including H2D/D2H transfer has
approximately $9.4\times$ headroom (with p99$-$min $<0.03$ ms, the variance is also
extremely small, and under sustained load the operating frequency remains
almost at the maximum clock owing to DVFS).

\begin{table}[t]
\centering
\caption{FP16 inference latency of EffNet-B0 on the Orin Nano (1000
iterations).}
\label{tab:orin_latency}
{\footnotesize\setlength{\tabcolsep}{4pt}
\begin{tabular}{l|r|r|r}\hline\hline
Metric & Mean & Median & p99 \\\hline
GPU compute (ms) & \textbf{3.139} & 3.138 & 3.150 \\
Incl.\ H2D/D2H (ms) & 3.192 & 3.192 & 3.204 \\
Throughput (images/s) & \multicolumn{3}{c}{\textbf{318} / 313 (GPU compute / incl.\ transfer)} \\\hline
\end{tabular}\par}
\end{table}

We confirmed there is no accuracy loss from FP16 by passing all test
images ($n{=}1{,}882$) through the real FP16 engine. The Orin FP16 accuracy
of 0.9267 and macro recall of 0.9354 agreed within error with the FP32
reference of onnxruntime on the Mac (0.9261 / 0.9351), and the per-image
argmax agreement was 99.95\% (1,881/1,882) (the single disagreement was a
case where FP16 flipped to correct). The about 0.2-point difference from
the PyTorch evaluation values in Table~\ref{tab:students_acc} (accuracy
0.9283, macro recall 0.9369) may stem from differences in the
preprocessing implementation (the PyTorch evaluation uses torchvision, the
real-hardware harness uses the ONNX-bundled preprocessing) and numerical
differences in the runtime (an experiment isolating the factor was not
performed), but is at least not an accuracy loss from FP16 itself (the
99.95\% FP16$\leftrightarrow$FP32 argmax agreement on the same engine
supports this). That is, \textbf{no substantial accuracy loss from FP16 was
observed}, and the accuracy of the deployed FP16 engine can be regarded as
equal to the FP32 measurement (unlike INT8 quantization, FP16 is
half-precision, not integer quantization). Note that the latency measured
on real hardware (3.19 ms/image including H2D/D2H, 3.14 ms/image GPU
compute) and the macro recall of 0.9354 are values for a \textbf{single
student EffNet-B0 (FP16)}. On the other hand, the highest accuracy of this
study, a macro recall of 0.9547, is an offline evaluation value of the
three-student ensemble (34.13M in total), and the two are separate
configurations. We measured on real hardware the latency of using this
highest-accuracy configuration at the edge (exclusive condition, same
trtexec settings). The FP16 inference latency (including H2D/D2H) of each
of the three students was 3.21 ms for EffNet-B0, 2.24 ms for MNV4, and
10.27 ms for ViT-Small, and the sum of these measured values is
\textbf{15.72 ms/image}. Because TensorRT 8.5.2 cannot optimize fused
attention (scaled dot-product attention) for ViT-Small and falls back to a
non-fused implementation, its latency is dominant, but even excluding
overhead such as softmax averaging, the three-student ensemble is
estimated to fit within the 30 ms budget (about $1.9\times$ headroom).

\subsection{Verification of INT8 and confirmation of FP16 deployment}
\label{sec:int8}

Because FP16 already exceeds the target by $9.4\times$ and no substantial
accuracy loss is observed, the motivation for INT8 is limited, but we
verified the potential for additional speed/memory savings on real
hardware via three routes. Evaluating all routes on the same full test set
($n{=}1{,}882$), results are shown in Table~\ref{tab:int8}. (a) PTQ of the
standard EffNet-B0 (Entropy2 calibration) shortened latency to 2.371 ms but
dropped accuracy to 0.8475 (a large accuracy drop from PTQ of the standard
EffNet-B0 was observed in this experiment). (b) QAT (quantization-aware
training) recovered the accuracy of 0.159 at the stage of only calibration
after Q/DQ insertion (before quantization-aware retraining) to 0.8895 by
retraining with quantization built in, but TensorRT 8.5.2 could not fully
fuse the Q/DQ nodes, making it slower than FP16 (GPU compute, fastest 5.969
ms) and its accuracy did not reach FP16. (c) The quantization-friendly
modified version (SE removal, SiLU$\to$ReLU6) had an FP32 ceiling of 0.88,
at or below FP16, and its PTQ on this TensorRT route dropped accuracy to
0.1530, nearly indistinguishable.

\begin{table}[t]
\centering
\caption{Three INT8 routes vs FP16 on the Orin.}
\label{tab:int8}
{\footnotesize
\begin{tabular}{l|r|r|r}\hline\hline
Configuration & Accuracy & Macro recall & GPU compute (ms) \\\hline
\textbf{FP16 (deployed)} & \textbf{0.9267} & \textbf{0.9354} & \textbf{3.139} \\
PTQ-INT8 & 0.8475 & 0.8569 & 2.371 \\
QAT-INT8 & 0.8895 & 0.8968 & 5.969 \\
Affinity-modified INT8 & 0.1530 & 0.1724 & 1.146 \\\hline
\end{tabular}\par}
{\footnotesize \hspace{1em}All routes evaluated on the same full test set
($n{=}1{,}882$).\par}
\end{table}

No INT8 route could \textbf{simultaneously exceed FP16 in both speed and
accuracy} (PTQ is faster than FP16 but drops accuracy substantially, and
QAT partially recovers accuracy but is slower than FP16 due to insufficient
fusion). Even with the newer TensorRT 10.7 (a relative comparison of GPU
compute alone via \texttt{trtexec} on the H200), QAT-INT8 is $1.66\times$
slower than FP16 due to unfused Q/DQ, so even with a version update the
QAT-INT8 of the standard EffNet-B0 does not exceed FP16. \textbf{Therefore,
we fixed the deployment at FP16.}

\subsection{Classification robustness at multiple resolutions assuming PTZ
magnification}
\label{sec:multires}

Although the deployed model's input is $224\times224$ pixels, in actual
operation the effective resolution of a cropped individual varies greatly
with the PTZ camera's zoom ratio and the distance to the target (the
lengthening of detection distance with larger turbines, mentioned in
Section~\ref{sec:related}, also promotes smaller and lower-resolution
imaged individuals). We therefore downscaled the $224$ central crop of the
test set ($n{=}1{,}882$) to 32/64/128 pixels once and re-upscaled it to
224, simulating the situation where an individual imaged at each native
resolution is stretched to 224 by PTZ, and evaluated the classification
performance of the deployed EffNet-B0 and the three-student ensemble
(Table~\ref{tab:multires}). At the 224-pixel equivalent, EffNet-B0
maintained a macro recall of 0.9369 (ensemble 0.9547), whereas at the
128-pixel equivalent it maintained 0.9117 (0.9445), and it dropped
substantially to 0.8164 (0.8791) at the 64-pixel equivalent and to 0.6442
(0.6529) at the 32-pixel equivalent. Therefore, the deployability claim of
this study is for inputs where the cropped individual occupies roughly 128
pixels square or more, and at long range the effective resolution must be
secured by PTZ zoom or detection-distance design. However, this evaluation
simulates low-resolution input by downscaling and re-upscaling 224-pixel
images and does not reproduce the atmospheric turbulence, video
compression, motion blur, and cropping error contained in actual
long-range PTZ images. Therefore, the values in Table~\ref{tab:multires}
are an upper-side guide that does not include these degradation factors and
are not a definitive guarantee of low-resolution performance in actual
operation. Note that the ensemble exceeded standalone EffNet-B0 at all
resolutions and kept its relative advantage even at low resolution.

\begin{table}[t]
\centering
\caption{Classification performance for inputs captured at each native
resolution and upscaled to 224 (test set, $n{=}1{,}882$).}
\label{tab:multires}
{\footnotesize
\begin{tabular}{r|r|r|r|r}\hline\hline
Native & \multicolumn{2}{c|}{EffNet-B0 (deployed)} & \multicolumn{2}{c}{Three-student ensemble} \\
(px) & Accuracy & Macro recall & Accuracy & Macro recall \\\hline
224 & 0.9283 & \textbf{0.9369} & 0.9506 & \textbf{0.9547} \\
128 & 0.9049 & 0.9117 & 0.9416 & 0.9445 \\
64  & 0.8071 & 0.8164 & 0.8698 & 0.8791 \\
32  & 0.6360 & 0.6442 & 0.6344 & 0.6529 \\\hline
\end{tabular}\par}
\end{table}

\subsection{Aggregation into the turbine-stop decision and rejection of
out-of-target species}
\label{sec:stopdecision}

We evaluate the operational characteristics when connecting this
classifier's output to the turbine-stop decision. In Japan, of these six
target species, five---Golden Eagle, White-tailed Eagle, Steller's Sea
Eagle, Mountain Hawk-Eagle (designated 1993), and Eastern Marsh Harrier
(designated 2017)---are designated as \textbf{Nationally Endangered
Species of Wild Fauna and Flora}~\cite{moekisho} under the Act on
Conservation of Endangered Species, and the Northern Goshawk was
de-designated in 2017. In particular, for the sea eagles (White-tailed and
Steller's Sea Eagle), the Ministry of the Environment provides specific
turbine-stop guidelines~\cite{moenv2022}. Globally, an operation that
targets species in higher IUCN threat categories (CR/EN, etc.) for
shutdown is generally assumed. However, designation as a Nationally
Endangered Species does not immediately mean it is subject to automatic
turbine shutdown; the actual stop criteria follow siting, permit
conditions, and the operator's operational rules. In this evaluation, as
an assumption for quantifying operational characteristics, we regard these
five species as stop targets and the Northern Goshawk as a non-stop
target. Under this assumption, aggregating the confusion matrix of the
deployed EffNet-B0 (six classes) into ``five stop-target species vs.\
Northern Goshawk'' gives an aggregated recall of 0.9885, an aggregated
precision of 0.9903, and a false-stop rate (erroneously stopping for a
non-target) of 0.0711 (16 of 225 Northern Goshawk images). The six-class
macro precision is 0.9309 and macro recall is 0.9369.

Because closed-set classification always assigns any input to one of the
six species even if it is outside the six targets, actual operation
requires \textbf{rejection of out-of-target species}. We therefore regarded
the maximum softmax probability (MSP) of the deployed EffNet-B0 as a
confidence and evaluated its ability to discriminate ID (the six target
species, test $n{=}1{,}882$) from OOD (1,455 images extracted from raptors
other than the six targets). The OOD set was constructed by extracting,
with an allow list, only the \textbf{97 species} taxonomically belonging to
the four orders Accipitriformes, Falconiformes, Cathartiformes, and
Strigiformes from the non-target species collected from iNaturalist, GBIF,
Wikimedia, etc.\ when building the V6 bird classifier (because extraction by
partial English-name match can admit non-raptors such as \textit{Western
Meadowlark} or \textit{Nighthawk}, we explicitly excluded these). Up to 20
images per species were randomly extracted with random seed 2026, for a
total of 1,455 images (duplicates removed at the file level). With MSP as
the ID/OOD discriminator, the AUROC was only 0.699 (ID MSP mean 0.85 vs.\
OOD 0.73), and at the confidence threshold $\tau{=}0.557$ that retains 95\%
of ID, the OOD rejection rate was 22.1\% (OOD false-accept rate FPR95 =
77.9\% at the same threshold). That is, \textbf{the ability to reject
out-of-target species by a simple confidence threshold is limited}, and to
connect a closed-set classifier to operation, it must be combined with
confidence calibration, dedicated open-set/OOD detection methods, and the
upstream bird/non-bird and rare-species detection stage (the parallel
study in Section~\ref{sec:vs_v6}). The deployability claim of this paper is
about the fine-grained identification of the six-species closed set, and
robust rejection of out-of-target species is future work.

\subsection{A/B of the teacher foundation-model generation (DINOv2-L vs.\
DINOv3-L)}
\label{sec:dinov3}

To evaluate the effect of the teacher-model choice on student performance,
in a controlled A/B varying only the teacher, we evaluated the
new-generation self-supervised foundation model DINOv3-L~\cite{simeoni}
(ViT-L/16, 300M). We fixed the data split, input, student settings,
$T/\alpha$, and epoch count exactly the same as the DINOv2-L teacher
version, and re-fine-tuned the teacher in fp32. The post-fine-tuning
teacher accuracies were close: macro recall 0.9789 (DINOv2-L) and 0.9758
(DINOv3-L). At a single seed the three-student ensemble was slightly higher
for DINOv2-L, but because the difference was small, we re-distilled with
each teacher over five student-distillation seeds ($\{42,43,44,45,46\}$)
and computed the mean$\pm$SD of the ensemble macro recall
(Table~\ref{tab:seedab}). The results were $0.9512\pm0.0031$ (DINOv2-L) and
$0.9533\pm0.0038$ (DINOv3-L). As in Section~\ref{sec:ceonly}, we analyze in
two ways with an explicit evaluation target. \textbf{(i)} The seed-paired
difference (v2$-$v3) was $-0.0021\pm0.0063$ over five seeds (an exploratory
comparison with $n{=}5$; DINOv2-L was superior in only one seed), and the
mean$\pm$SD range included zero. \textbf{(ii)} Separately, an image-level
paired bootstrap ($B{=}2000$, uncertainty of the test sample) of a single
ensemble integrating the five seeds by probability averaging gave a
difference of $0.0000$, a 95\% confidence interval of $[-0.0048, +0.0049]$,
and a fraction of resamples in which v2 exceeded v3 of 0.49 (not a
$p$-value). In both analyses, \textbf{the difference between the two teacher
generations is not recognized, within seed variation and sample
uncertainty}. No clear improvement from changing the teacher generation was
confirmed, and \textbf{we decided to keep the already-verified, operationally
proven DINOv2-L as the teacher} (a model should not be replaced on the sole
basis of a newer generation).

\begin{table}[t]
\centering
\caption{Teacher-generation A/B over five distillation seeds
(three-student ensemble macro recall, mean$\pm$SD; new test set,
$n{=}1{,}882$).}
\label{tab:seedab}
{\footnotesize
\begin{tabular}{l|c|c|p{58mm}}\hline\hline
Teacher & Mean$\pm$SD & Range & Paired difference / test \\\hline
DINOv2-L & $0.9512\pm0.0031$ & 0.9467--0.9547 & (i) seed-paired diff v2$-$v3 $=-0.0021\pm0.0063$ ($n{=}5$) \\
DINOv3-L & $\mathbf{0.9533\pm0.0038}$ & 0.9466--0.9555 & (ii) seed-integrated bootstrap 95\% CI $[-0.005,+0.005]$, v2-superior fraction 0.49 \\\hline
\end{tabular}\par}
\end{table}

\section{Discussion}
\label{sec:discussion}

\subsection{Inter-species boundary blurring caused by rare-class scarcity}

Before expansion, the Steller's Sea Eagle training count was only 324
images (just under 8\% of the training set), markedly fewer than the
related White-tailed Eagle. In the process of increasing the Steller's Sea
Eagle training count to 1,432 images (16\%) after expansion, the
rate of misclassification of White-tailed Eagle as Steller's Sea Eagle
decreased from 61\% to 15\% of errors. Because teacher retraining and
student re-distillation proceed in parallel in this process, it is not a
strict causal separation, but it is consistent with the typical case that
``even when the data count itself is sufficient, recall drops if the
related-species side is thin,'' suggesting that a design that
preferentially expands the rare-class side after grasping the
genus/subfamily-level distribution of the classification target set may be
effective. Discriminating related species is a typical difficulty of FGVC
represented by CUB-200-2011~\cite{wah}, and this result is a concrete
example showing that, in addition to the design of the loss function, the
distribution of the training data (especially at the genus/subfamily-level
granularity) greatly affects FGVC accuracy.

\subsection{Effectiveness of re-fine-tuning the teacher}

Because of the expansion of the data distribution, the old teacher's
evaluation value dropped from 0.9706 on the old test set to 0.8985 on the
new test set (the two are test sets from different populations, so this is a
change in the evaluation distribution, not degradation of the model
itself). If distillation continued in this state, the student's performance
would plateau, but re-fine-tuning the teacher (11 minutes) improved the
student's accuracy by up to +3.24 points (ViT-Small in
Table~\ref{tab:students_acc}, old-teacher-logit version
$0.9102\rightarrow$ new-teacher-logit version 0.9426), a clearly positive
return on investment. This shows the effectiveness of the iterative loop
``data expansion $\to$ teacher retraining $\to$ student re-distillation.''

\subsection{Choosing between FP16 and INT8 in edge deployment}

On Orin Nano-class edge GPUs, FP16 has no accuracy loss and greatly
exceeds the target latency, so the additional benefit of INT8 in this task
is limited. INT8 drops accuracy substantially with PTQ (the quantization
sensitivity of the standard EffNet-B0's Squeeze-Excite~\cite{hu} and SiLU),
and while QAT recovers accuracy, it is slower than FP16 due to the
insufficient Q/DQ fusion of TensorRT 8.5.2. In edge implementation, it is
sound to first confirm ``whether FP16 fits within the target latency'' and,
as long as it does, make FP16 the first choice. INT8 is justified only when
there are stricter latency/power constraints, or when a newer runtime with
improved Q/DQ fusion (the TensorRT 10 series) and a quantization-friendly
backbone are both available.

\subsection{A newer teacher-model generation does not necessarily bring a
performance improvement}

The choice of teacher model can affect student performance, but in this
experiment the teacher accuracies of DINOv2-L and DINOv3-L (0.9789 vs.\
0.9758) and the distilled student-ensemble accuracies (five-seed mean
$0.9512\pm0.0031$ vs.\ $0.9533\pm0.0038$) were both close, and no clear
performance improvement from changing the teacher generation was confirmed
(Section~\ref{sec:dinov3}). Rather, the teacher accuracy was higher for
DINOv2-L while the student mean accuracy was slightly higher for DINOv3-L,
inverting the ordering of the two. That is, student accuracy cannot be
predicted monotonically from the teacher accuracy alone. Because the
distillation loss also includes the hard-label cross-entropy
(Eq.~\ref{eq:kd}), the student performance is not mathematically upper-bounded
by the teacher performance either. Therefore, even for a new-generation
general-purpose foundation model, adoption should be judged only after
measuring the spillover to the student across seeds in the target domain,
and a teacher should not be replaced on the sole basis of a newer
generation, as this A/B shows.

\subsection{Contribution of knowledge distillation and rigor of the data
split}

The five-seed control against the non-distillation (CE-only) baseline
(Section~\ref{sec:ceonly}) showed that, for the three-student ensemble in
this task, the difference between knowledge distillation and
non-distillation is within seed variation and no clear ensemble improvement
from distillation is confirmed. Even for the single deployed EffNet-B0, the
distillation improvement was limited to a single seed and vanished in the
five-seed mean (paired difference $+0.0017\pm0.0156$), and at the ensemble
level non-distillation was rather slightly superior. That is, the primary
accuracy improvement in this task stems not from knowledge distillation
itself but from rare-class data expansion and teacher re-fine-tuning.
Distillation is a valid lightweight-conversion means that leverages teacher
knowledge, but under this setting and model scale its additional
contribution to ensemble accuracy is limited, which may depend on the
distillation temperature/weight and the representational-capacity gap
between teacher and student. Also, the fact that the ensemble macro recall
dropped only about 2.2 points, from 0.9547 to 0.9326, under the
video/source-image-level group split (Section~\ref{sec:group_split})
supports that the source leakage feared with an image-level split has a
limited effect (at least at the video/source-image granularity) and that
the improvement from data expansion is preserved even under conditions that
mitigate leakage.

\subsection{Limitations and future work}

Because this study targeted six species across four subfamilies, the effect
of hierarchical aggregation based on subfamily taxonomy was limited.
Re-evaluation on a deeper taxonomy of 60--100 species is future work. In
this paper, we measured the real-hardware latency of the single student
EffNet-B0 (FP16) as the deployment configuration and separately measured
the FP16 latency of the three students (sum 15.72 ms/image), but an
integrated measurement as a continuous execution system combining them was
not performed; measuring the latency, power, and thermals of continuously
running the three-student ensemble on real hardware, and demonstrating
end-to-end real-time operation coupled with a detection/tracking pipeline
from a fixed camera, is the next stage of work. Also, to further raise the
performance ceiling of the teacher model, evaluation of larger DINOv3
variants (ViT-H+/7B) remains.

\section{Conclusion}
\label{sec:conclusion}

To support the avoidance of raptor bird strikes at wind-power facilities on
edge devices, we designed a lightweight distillation system with DINOv2-L
as the teacher, and through a combination of rare-class reinforcement via
video-frame extraction and teacher re-fine-tuning, the three-student
ensemble achieved a macro recall of $0.935\pm0.004$ (five
student-distillation seeds; 0.955 on the standard image-level split) under
a group split that mitigates source leakage at the video/source-image
level, using a configuration with approximately $1/8.9$ of the teacher's
parameters. We reduced related-species confusion from 61\% to 15\% on a
direct metric. We built a TensorRT FP16 engine of the deployed student
EffNet-B0 on an NVIDIA Jetson Orin Nano and achieved, as a single student,
3.19 ms/image including H2D/D2H transfer (about 313 images/s,
approximately $9.4\times$ headroom against the 30 ms target) on real
hardware with no
substantial accuracy loss relative to FP32. For INT8, none of PTQ, QAT, or
quantization-friendly modification simultaneously exceeded FP16 in both
speed and accuracy, so we fixed the deployment at FP16. Furthermore, we
performed a controlled A/B (five seeds) replacing the teacher with
DINOv3-L, but the difference in student-ensemble accuracy was within seed
variation, and we kept the operationally proven DINOv2-L teacher. This
approach may be extended to edge-based species identification of other
raptor groups and rare birds. Future work will focus on broader
generalization by expanding the number of target species and on
demonstrating fully end-to-end real-time operation.

\section*{Acknowledgments}

As computing resources, we used the M system and HPCI-AISS1 of the
Foundation for Computational Science (FOCUS). In data collection, we thank
the contributors and providers who publish observation information and
images on iNaturalist, GBIF, Wikipedia, and Wikimedia Commons; we used the
images and metadata in accordance with each data's license conditions. We
thank Eurus Energy Holdings Corporation for their cooperation in providing
part of the bird photographs and flight videos obtained from May 2020 to
November 2024. We also thank the parties who advised on still-image
extraction from videos and species identification.



\begin{thebibliography}{99}

\bibitem{aschwanden}
Aschwanden, J., Stark, H., Peter, D. et al.:
Bird collisions at wind turbines in a mountainous area related to bird movement intensities measured by radar,
{\it Biol.\ Conserv.}, Vol.\,220, pp.\,228--236 (2018).

\bibitem{marques}
Marques, A.\,T., Batalha, H., Rodrigues, S. et al.:
Understanding bird collisions at wind farms: An updated review on the causes and possible mitigation strategies,
{\it Biol.\ Conserv.}, Vol.\,179, pp.\,40--52 (2014).

\bibitem{may}
May, R., Reitan, O., Bevanger, K. et al.:
Mitigating wind-turbine induced avian mortality: Sensory, aerodynamic and cognitive constraints and options,
{\it Renew.\ Sustain.\ Energy Rev.}, Vol.\,42, pp.\,170--181 (2015).

\bibitem{shiraki2012}
Shiraki, S.:
Current status of collisions of White-tailed Eagles (\textit{Haliaeetus albicilla}) with wind turbines in Hokkaido,
{\it Japanese Journal of Conservation Ecology}, Vol.\,17, No.\,1, pp.\,85--96 (2012) (in Japanese).

\bibitem{hinton}
Hinton, G., Vinyals, O. and Dean, J.:
Distilling the knowledge in a neural network,
arXiv: 1503.02531 (2015).

\bibitem{oquab}
Oquab, M., Darcet, T., Moutakanni, T. et al.:
DINOv2: Learning robust visual features without supervision,
{\it Trans.\ Mach.\ Learn.\ Res.}, January (2024).

\bibitem{ueda2010}
Ueda, M., Fukuda, Y. and Takada, R.:
Differences in flight behavior between White-tailed Eagles and Steller's Sea Eagles,
{\it Bird Research}, Vol.\,6, pp.\,A43--A52 (2010) (in Japanese).

\bibitem{maki2012}
Maki, K.:
Identification Guide to Eagles, Hawks and Falcons,
Heibonsha, Tokyo (2012) (in Japanese).

\bibitem{simeoni}
Sim\'eoni, O., Vo, H.\,V., Seitzer, M. et al.:
DINOv3,
{\it Trans.\ Mach.\ Learn.\ Res.} (2026), arXiv:2508.10104.

\bibitem{gou}
Gou, J., Yu, B., Maybank, S.\,J. and Tao, D.:
Knowledge distillation: A survey,
{\it Int.\ J.\ Comput.\ Vis.}, Vol.\,129, pp.\,1789--1819 (2021).

\bibitem{touvron}
Touvron, H., Cord, M., Douze, M. et al.:
Training data-efficient image transformers \& distillation through attention,
{\it Proc.\ Int.\ Conf.\ Mach.\ Learn.\ (ICML)}, PMLR 139, pp.\,10347--10357 (2021).

\bibitem{wah}
Wah, C., Branson, S., Welinder, P., Perona, P. and Belongie, S.:
The Caltech-UCSD Birds-200-2011 dataset,
Technical Report CNS-TR-2011-001, California Institute of Technology (2011).

\bibitem{vanhorn}
Van Horn, G., Mac Aodha, O., Song, Y. et al.:
The iNaturalist species classification and detection dataset,
{\it Proc.\ IEEE Conf.\ Comput.\ Vis.\ Pattern Recognit.\ (CVPR)}, pp.\,8769--8778 (2018).

\bibitem{wightman}
Wightman, R.:
PyTorch Image Models (timm) (online),
available at \url{https://github.com/huggingface/pytorch-image-models} (accessed 2026-05-05).

\bibitem{qin}
Qin, D., Leichner, C., Delakis, M. et al.:
MobileNetV4 -- Universal models for the mobile ecosystem,
{\it Proc.\ Eur.\ Conf.\ Comput.\ Vis.\ (ECCV)}, LNCS, Vol.\,15098, pp.\,78--96 (2024), DOI:10.1007/978-3-031-73661-2\_5.

\bibitem{dosovitskiy}
Dosovitskiy, A., Beyer, L., Kolesnikov, A. et al.:
An image is worth 16x16 words: Transformers for image recognition at scale,
{\it Proc.\ Int.\ Conf.\ Learn.\ Represent.\ (ICLR)} (2021).

\bibitem{tan}
Tan, M. and Le, Q.:
EfficientNet: Rethinking model scaling for convolutional neural networks,
{\it Proc.\ Int.\ Conf.\ Mach.\ Learn.\ (ICML)}, pp.\,6105--6114 (2019).

\bibitem{tensorrt}
NVIDIA Corporation:
NVIDIA TensorRT (online),
available at \url{https://developer.nvidia.com/tensorrt} (accessed 2026-07-10).

\bibitem{jacob}
Jacob, B., Kligys, S., Chen, B. et al.:
Quantization and training of neural networks for efficient integer-arithmetic-only inference,
{\it Proc.\ IEEE Conf.\ Comput.\ Vis.\ Pattern Recognit.\ (CVPR)},
pp.\,2704--2713 (2018).

\bibitem{gholami}
Gholami, A., Kim, S., Dong, Z. et al.:
A survey of quantization methods for efficient neural network inference,
arXiv: 2103.13630 (2021).

\bibitem{hu}
Hu, J., Shen, L. and Sun, G.:
Squeeze-and-Excitation networks,
{\it Proc.\ IEEE Conf.\ Comput.\ Vis.\ Pattern Recognit.\ (CVPR)},
pp.\,7132--7141 (2018).

\bibitem{moenv2011}
Ministry of the Environment, Nature Conservation Bureau:
Guidelines for the appropriate siting of wind-power facilities with respect to birds and other wildlife,
Ministry of the Environment, Tokyo (2011) (in Japanese).

\bibitem{moenv2022}
Ministry of the Environment, Nature Conservation Bureau:
Guidelines for examining and implementing bird-strike prevention measures for sea eagles at wind-power facilities (revised August 2022),
Ministry of the Environment, Tokyo (2022) (in Japanese).

\bibitem{yoshihashi2015}
Yoshihashi, R., Kawakami, R., Iida, M. and Naemura, T.:
Construction of a bird image dataset for ecological investigations,
{\it Proc.\ IEEE International Conference on Image Processing (ICIP)}, pp.\,4248--4252 (2015).

\bibitem{yoshihashi}
Yoshihashi, R., Kawakami, R., Iida, M. and Naemura, T.:
Bird detection and species classification with time-lapse images around a wind farm: Dataset construction and evaluation,
{\it Wind Energy}, Vol.\,20, No.\,12, pp.\,1983--1995 (2017).

\bibitem{mcclure2021}
McClure, C.\,J.\,W., Rolek, B.\,W., Dunn, L. et al.:
Eagle fatalities are reduced by automated curtailment of wind turbines,
{\it J.\ Appl.\ Ecol.}, Vol.\,58, No.\,3, pp.\,446--452 (2021).

\bibitem{mcclure2022}
McClure, C.\,J.\,W., Rolek, B.\,W., Dunn, L. et al.:
Confirmation that eagle fatalities can be reduced by automated curtailment of wind turbines,
{\it Ecol.\ Solut.\ Evidence}, Vol.\,3, No.\,3, e12173 (2022).

\bibitem{iucn}
IUCN:
The IUCN Red List of Threatened Species (online),
available at \url{https://www.iucnredlist.org/} (accessed 2026-07-10).

\bibitem{inat2024}
iNaturalist contributors, iNaturalist:
iNaturalist Research-grade Observations,
GBIF Occurrence dataset (online),
available at \url{https://doi.org/10.15468/ab3s5x} (accessed 2026-05-04).

\bibitem{gbif}
GBIF.org:
The Global Biodiversity Information Facility (online),
available at \url{https://www.gbif.org/} (accessed 2026-05-04).

\bibitem{gemini}
Google:
Gemini API (online),
available at \url{https://ai.google.dev/gemini-api/docs} (accessed 2026-02-15).

\bibitem{torchddp}
Li, S., Zhao, Y., Varma, R. et al.:
PyTorch distributed: experiences on accelerating data parallel training,
{\it Proc.\ VLDB Endow.}, Vol.\,13, pp.\,3005--3018 (2020).

\bibitem{onnx}
Bai, J., Lu, F., Zhang, K. et al.:
ONNX: Open Neural Network Exchange (online),
available at \url{https://github.com/onnx/onnx} (accessed 2026-05-05).

\bibitem{modelopt}
NVIDIA Corporation:
NVIDIA TensorRT Model Optimizer (online),
available at \url{https://github.com/NVIDIA/TensorRT-Model-Optimizer} (accessed 2026-07-10).

\bibitem{focus}
Foundation for Computational Science:
FOCUS Supercomputer User Guide (online),
available at \url{https://www.j-focus.jp/user_guide/} (accessed 2026-07-10) (in Japanese).

\bibitem{h200}
NVIDIA Corporation:
NVIDIA H200 Tensor Core GPU (online),
available at \url{https://www.nvidia.com/en-us/data-center/h200/} (accessed 2026-07-10).

\bibitem{orin}
NVIDIA Corporation:
NVIDIA Jetson Orin Nano (online),
available at \url{https://www.nvidia.com/en-us/autonomous-machines/embedded-systems/jetson-orin/} (accessed 2026-07-10).

\bibitem{m2max}
Apple Inc.:
MacBook Pro (14-inch, 2023) Technical Specifications (online),
available at \url{https://support.apple.com/en-us/111340} (accessed 2026-07-10).

\bibitem{moekisho}
Ministry of the Environment:
List of Nationally Endangered Species of Wild Fauna and Flora (online),
available at \url{https://www.env.go.jp/nature/kisho/domestic/list.html} (accessed 2026-07-25) (in Japanese).

\end{thebibliography}
\end{document}